%% file: acl_latex.tex
\pdfoutput=1
\documentclass[11pt]{article}
\usepackage[final]{acl}
\usepackage{times}
\usepackage{latexsym}
\usepackage[T1]{fontenc}
\usepackage[utf8]{inputenc}
\usepackage{microtype}
\usepackage{inconsolata}
\usepackage{soul}
\definecolor{dark green}{RGB}{30,130,50}
\usepackage[normalem]{ulem}
\useunder{\uline}{\ul}{}
\usepackage{tabularx}
\usepackage{graphicx}
\newcommand{\data}{\texttt{PUMA}}
\newcommand{\model}{\texttt{PLASMA}}
\usepackage{enumitem}
\usepackage{soul}
\usepackage{colortbl}
\usepackage{graphicx}
\usepackage{lipsum}
\usepackage{multirow}
\usepackage{booktabs}
\usepackage{arydshln}
\usepackage{caption}
\usepackage{hyperref}
\usepackage{tabularray}
\usepackage{amsmath,amssymb}
\UseTblrLibrary{booktabs}
\usepackage{caption}
\usepackage{ragged2e}
\usepackage{tabularx}

\title{\textit{No perspective, no perception!!} Perspective-aware Healthcare Answer Summarization}
\author{
  Gauri Naik$^{1}$, \
  Sharad Chandakacherla$^{2}$, \ 
  Shweta Yadav$^{2}$, \ 
  Md. Shad Akhtar$^{1}$ \\
  $^{1}$IIIT Delhi,
   $^{2}$University of Illinois at Chicago \\
  \texttt{\{gaurin, shad.akhtar\}@iiitd.ac.in}, 
  \texttt{\{schand65, shwetay\}@uic.edu.in}
}
\begin{document}

\maketitle
\begin{abstract}
Healthcare Community Question Answering (CQA) forums offer an accessible platform for individuals seeking information on various healthcare-related topics. 
People find such platforms suitable for self-disclosure, seeking medical opinions, finding simplified explanations for their medical conditions, and answering others' questions. However, answers on these forums are typically diverse and prone to off-topic discussions.
It can be challenging for readers to sift through numerous answers and extract meaningful insights, making answer summarization a crucial task for CQA forums. While several efforts have been made to summarize the community answers, most of them are limited to the open domain and overlook the different perspectives offered by these answers. 
To address this problem, this paper proposes a novel task of perspective-specific answer summarization. 
We identify various perspectives, within healthcare-related responses and frame a perspective-driven abstractive summary covering all responses. To achieve this, we annotate $3167$ CQA threads with $6193$ perspective-aware summaries in our \data\ dataset. Further, we propose \model, a prompt-driven controllable summarization model. To encapsulate the perspective-specific conditions, we design an energy-controlled loss function for the optimization. We also leverage the prefix tuner to learn the intricacies of the health-care perspective summarization.
Our evaluation against five baselines suggests the superior performance of \model\ by a margin of \textasciitilde$1.5-21\%$ improvement. We supplement our experiments with ablation and qualitative analysis. 
\end{abstract}

\section{Introduction}


In this digital age, community question-answering (CQA) platforms like Quora, Reddit, and Yahoo! Answers have significantly transformed the way we exchange information. These platforms enable users from around the globe to share knowledge, experiences, and opinions, fostering a unique collaborative environment. Among the various topics discussed on these platforms, medical advising and interactions have gained notable popularity, such as Reddit's \href{https://www.reddit.com/r/AskDocs/}{\texttt{r/AskDocs}}. Users seek help by posting their questions and peers respond to them. However, the diverse nature of responses, as well as their overwhelming number, make finding reliable medical insights a challenging and non-trivial task. 
Summarizing the responses (or answers) in a concise and meaningful way offers a tangible solution. Moreover, these responses offer a wide range of perspectives such as \textit{personal experiences}, \textit{factual information}, \textit{advice}, etc., that an end user finds relevant \cite{fabbri-etal-2022-answersumm, chan2012community, chowdhury2019cqasumm}.
For example, Figure \ref{tab:table1} depicts an instance of a CQA thread where a user seeks advice on alternatives to surgery for gallstones. In response, peers responded with varied perspectives, e.g., peer1 provided perspectives on \textit{general information}, offer some \textit{suggestion}, talks about personal \textit{experience} and potential \textit{implications}. We observe a similar trend in other responses as well, where users present perspectives on the causes of the medical problem or pose questions to better understand the context or situation presented in the main question.
Capturing these diverse \textit{perspectives} in summaries is crucial \citep{fabbri-etal-2022-answersumm,chaturvedi2024aspect} since these varied insights are invaluable for users to make informed healthcare decisions and access appropriate support.


\begin{table*}[]
    \centering
    \renewcommand*{\arraystretch}{1.2}%
    \resizebox{\textwidth}{!}{
    \begin{tabular}{l|l|p{26cm}} \hline
         \rowcolor{magenta!60!white} \multicolumn{3}{l}{Question: I was just diagnosed with gallstones in my gall bladder I really don’t want to have surgery and have been told that there are other ways to get rid of the stones. Suggestions?} \\ \hline
         & Answer 1 & \textcolor{blue}{Most gallstones are made of pure cholesterol.} You might \textcolor{dark green}{try a diet with low fat and very low saturated fats}. \textcolor{blue}{Reducing the saturated fats will lower blood cholesterol and may (I'm not promising anything!) make the stones smaller}. \textcolor{dark green}{Lowering your total fat intake may also help reduce or prevent pain}. \textcolor{blue}{Gallstones hurt because when you eat fat, the stomach senses it and tells the gallbladder to release some of the gall into the intestines to help digest the fat. If you have stones, that squeezing is about like squeezing a hand full of pointy rocks}. However, \textcolor{red}{I've had the surgery, and it really isn't a big deal}. \textcolor{blue}{There is minimal scarring}, …. \textcolor{brown}{If you leave the gallstones there, they can get large enough to damage the gallbladder, with the result of a bad infection and that can cause death}, which is a very Bad Thing!" \\ \cline{2-3}
         & Answer 2 & \textcolor{violet}{Have you seen a gastroenterologist?} \textcolor{blue}{They can do a minimally invasive procedure called an ERCP (Endoscopic Retrograde Cholangiopancreatography. An ERCP won't get rid of the stones that are in the gallbladder...just the stones that are stuck in the duct …. They can make a tiny cut in the duct and pull gallstones out with a small balloon. If the stone is too large, they have equipment that will crush the stone so it will pass freely}. \textcolor{red}{I had the surgery myself about 10 years ago. It's not as bad as you'd imagine, and you feel much better after it's over. You might still have phantom pain for a while, but it's nowhere near as bad as the pain you started with}. \textcolor{dark green}{A diet high in fat will make gallbladder disease worse, but you can't really get rid of the stones unless they pass naturally or you have them removed, either in surgery or with an ERCP}. \\ \cline{2-3}
         & Answer 3 & \textcolor{dark green}{The best remedy is surgery}. \textcolor{red}{I had surgery to have kidney stones removed. The surgery isn't as bad as you think it may be}. \\  \hline
         \rowcolor{green!10!orange}\multicolumn{3}{c}{\textcolor{white}{\bf Perspective-based summaries}} \\  \hline
         & \textcolor{blue}{Information} & Reducing saturated fats may shrink gallstones as they’re mostly made of cholesterol. Gallstone pain occurs when the gallbladder squeezes to aid digestion on fat consumption. An ERCP procedure by a gastroenterologist can remove stones stuck in the duct leading to the intestine. This minimally invasive technique involves extracting stones or crushing larger ones for easier passage, but it doesn't eliminate stones within the gallbladder itself. \\ \cline{2-3}
         & \textcolor{brown}{Cause} & Gallstones left untreated can harm the gallbladder, causing severe infection and potentially death. \\ \cline{2-3}
         & \textcolor{dark green}{Suggestion} & To eliminate gallstones without surgery, a low-fat diet, particularly low in saturated fats, as it may help reduce pain associated with gallbladder disease. Ultimately, surgical or medical intervention like ERCP may be necessary for complete removal if stones don’t pass naturally. \\ \cline{2-3}
         & \textcolor{red}{Experience} & Multiple people shared their experience of undergoing surgery to remove kidney stones, assuring that the procedure wasn't as daunting as expected. Despite the possibility of post-operative discomfort, the relief from the original pain was significant. \\ \cline{2-3}
         & \textcolor{violet}{Question} & It was asked if the person had seen a gastroenterologist \\ \bottomrule
    \end{tabular}}
    \caption{An example from the \data\  that illustrates the idea behind \textit{Perspectives}. 
\textcolor{blue}{Blue}: Information, \textcolor{red}{Red}: Experience, \textcolor{violet}{Violet}: Question, \textcolor{brown}{Brown}: Cause, \textcolor{dark green}{Green}: Suggestion. The color-coded spans are grouped and then used to write abstractive summaries. The summaries are marked with their perspective's corresponding color. Best viewed in color.}
    \label{tab:table1}
\end{table*}


Despite the apparent need, existing research on medical summarization predominantly focuses on medical reports \cite{michalopoulos-etal-2022-medicalsum} or medical dialogues \cite{joshi-etal-2020-dr}, and overlooks the role of perspectives in crafting effective summaries. Recently, \cite{bhattacharya2022lchqa,chaturvedi2024aspect} identified the role of perspective-guided summarization and created a dataset with $200$ CQA threads labeled with various perspectives. While it is a novel effort in this direction, the small dataset size limits the generalizability of the findings and the potential for training machine learning models.   



Considering these research gaps, we propose a novel perspective-specific answer summarization task in a CQA setup. 
Given a CQA thread (a question $Q$ and a set of answers $A$) and a desired perspective $P$, we aim to generate a concise summary $S^P$ that reflects the perspective $P$ across all answers. To achieve this, we build a novel perspective-aware summary annotated corpus of medical question-answers, \data\footnote{\textbf{P}erspective s\textbf{UM}marization d\textbf{A}taset}, which comprises $3167$ CQA threads with $\sim10K$ answers. Each answer in \data\ is annotated with five perspectives, i.e., \textit{`cause'}, \textit{`suggestion'}, \textit{`experience'}, \textit{`question'}, and \textit{`information'}, motivated by the work of \newcite{bhattacharya2022lchqa}. Consequently, we manually annotate a concise and relevant summary for each perspective -- each CQA thread has at most five perspective-specific summaries. 


Subsequently, we introduce \model\footnote{\textbf{P}erspective-aware hea\textbf{L}thcare \textbf{A}nswer \textbf{S}u\textbf{M}mariz\textbf{A}tion}, a novel energy-optimized transformer-based model for controllable perspective-guided summary generation. It aims to encapsulate essential information from answers and also reflect the attributes/perspectives, such as embodying a personalized tone and/or structure, in their generated summaries. To incorporate multiple attributes in the generation, we devise a prompt-learning-based strategy, where for each control attribute, we prepend the description of control attributes to the input source as hard prompts and also assign a set of trainable parameters called prefixes to our foundational model (i.e., Flan-T5).




Due to the conceptual nature of the control attributes, it is often challenging to assess and enforce the constraint on the generated summary with only the prompt-based strategy \cite{liu2021dexperts,yang2021fudge}. To properly enforce the constraints, we develop an energy-controlled objective function that computes the energy values separately for each attribute and enforces their inclusion in the generated summary. It forms a linear combination of multiple energy values to obtain a distribution whose samples satisfy all the attributes/constraints of the summary generation task.

We benchmark \model\ against five comparative systems and report the performances across ROUGE, Meteor, BERTScore, and BLEU. Our findings state that our model achieves superior performance across all metrics with a remarkable improvement of \textasciitilde $1.5-21.8\%$ compared to the closest baseline. We further complement our experiments with qualitative analysis against the best baseline. 

\paragraph{Contributions:} Our contributions are summarized below: 

\begin{itemize}[leftmargin=*, noitemsep]
   \item We develop a perspective-aware answer summarization dataset, \data, within the healthcare domain comprising of $3167$ CQA threads annotated with five domain-centric \textit{perspectives}.


    \item We design a novel prompt-based controllable text summarization model, \model. It combines prefix tuning with a perspective-specific energy-controlled loss function to enforce the controlling parameters in the generated summary.
    
    \item We evaluate our model against five baselines to verify significant improvements. Additionally, we also report thorough qualitative and quantitative analysis along with the ablation studies, to further validate our findings.

\end{itemize}

\noindent The \model\ model and the \data\ dataset are available at \url{https://github.com/GauriNaik826/PUMA-PLASMA-ACL}.

\section{Related Work}
Recent advancements in pre-trained language models (PLMs) have markedly improved performance in abstractive text summarization tasks. Notable examples include BART \cite{lewis2019bart}, T5 \cite{raffel2019exploring}, and PEGASUS \cite{zhang2019pegasus}, which have achieved state-of-the-art results particularly in summarizing news content, as demonstrated on large datasets such as CNN/DailyMail \cite{hermann2015teaching} and XSum \cite{narayan-etal-2018-dont} \cite{huang-etal-2023-summaries,chen2021capturing}.

In the biomedical and healthcare domain, significant advancements have been made in summarizing diverse types of content, including biomedical literature \cite{soleimani-etal-2022-zero}, consumer healthcare questions \cite{yadav2022chqsumm,yadav2022towards,yadav2023towards,yadav2021reinforcement,yadav2022question,savery2020question}, and medical notes \cite{hsu-etal-2020-characterizing}. These efforts predominantly utilize pre-trained language models (PLMs) such as BioBERT \cite{lee2020biobert}, BioBART \cite{yuan2022biobart}, and clinicalBERT \cite{huang2020clinicalbert}, which have been trained on extensive biomedical corpora like PubMed and MIMIC-III. Although these models demonstrate remarkable proficiency in generating fluent summaries, they often fall short in producing faithful summaries.



Early research in multi-document summarization (MDS), like  \citet{liu2018generating}, focused on extracting key information across documents and produce a unified summary. A similar idea is underlined in \citet{fabbri-etal-2019-multi}, for the news domain CNN/Daily Mail corpus \cite{hermann2015teaching}. Extensive research on news articles was presented as a part of the DUC\footnote{\url{http://duc.nist.gov/}} and TAC\footnote{\url{https://tac.nist.gov/}} tasks.
\citet{fabbri2021multi} introduced a query-focused multi-perspective summarization on a QA dataset with sentence-level spans. \citet{joshi-etal-2020-dr} and \citet{michalopoulos2022medicalsum} do the same by exploiting local and global features of the text. CTRLsum \cite{he2020ctrlsum} introduces a method that allows interaction during inference without predefined aspects to guide the model. CQASumm \cite{chowdhury2019cqasumm} highlighted the challenges of applying MDS on high-variance, opinion-based CQA data, revealing the limitations of modelling on fact-rich data. The dataset released by \citet{savery2020question} is the first in the medical domain to evaluate query-focused summaries albeit using only managed sources for data.

Perspective-based summarization is typically a two-step process of first identifying relevant sentences, followed by summarization. In AnswerSumm \cite{fabbri-etal-2022-answersumm, fabbri2021multi}, 
a model is used to extract sentences similar to the query. Alternatively, SpanBERT \citep{joshi-etal-2020-spanbert}, with its modified pre-training process, has shown good results in span-related tasks. In the biomedical realm, \citep{abaho-etal-2021-detect} used both word-level and sentence-level attention to detect medical outcome spans. The work \citep{ghosh2022span} uses dependency trees with a GCN \cite{zhang-etal-2018-graph} and a transformer to detect spans with \textit{dysfluencies}.

Our work here is unique in these ways: \textbf{(1)} we work on multi-document summarization on healthcare CQA data, hence not limiting ourselves to managed medical sources 
and \textbf{(2)} we employ phrase-level annotations to capture dense information. 



\section{Dataset}
\label{sssec:ann_scheme}
This section describes the dataset development process of \data, which is comprised of two stages of annotations: a) perspective and span identification; and b) perspective-driven summary. 

\subsection{Data Collection and Preprocessing}
We begin by collecting samples from the \textbf{L6 - Yahoo! Answers CQA}\footnote{L6 - Yahoo! Answers Comprehensive Questions and Answers, shared under Yahoo! Webscope program for research purposes \url{https://webscope.sandbox.yahoo.com/catalog.php?datatype=l&did=11} \label{footnote_1}.}. It is a large-scale dataset extracted from the Yahoo Answers forum, consisting of records until October 2007. We filtered the dataset on the healthcare category and randomly selected $10000$ questions with upto 10 answers each. The filtered records span a variety of topics including `\textit{Diabetes}', `\textit{Dental}', `\textit{Cancer}', etc (\S \ref{subsec:categories}). 

\begin{figure*}[ht]
    \centering
        \includegraphics[width=1\linewidth]{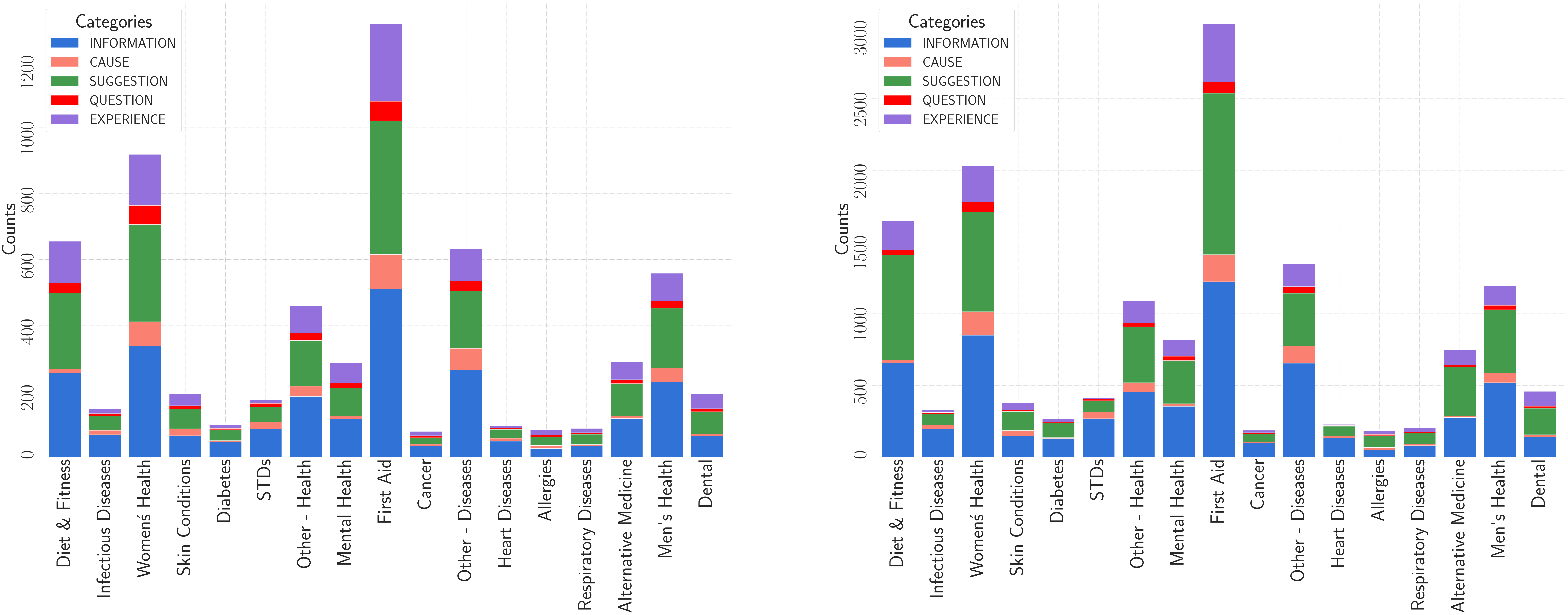}
        \caption{Class-wise distribution of spans (left) and summaries(right) across health categories.}
        \label{fig:fig_cat_sum_span_dist}
\end{figure*}

\subsection{Annotation Guideline} \label{annotation_process}
In response to a question, each user can respond in a different way and with varied perspectives. We drew inspiration from \cite{bhattacharya2022lchqa} to identify domain specific perspectives by examining multiple documents. Initially, we identified the following 7 perspectives: \textit{`cause'}, `\textit{suggestion}', `\textit{information}', `\textit{question}', `\textit{experience}', `\textit{clarification}', and `\textit{treatment}'. After careful consideration, we merge the \textit{`clarification'} perspective into the \textit{`information'} perspective due to their overlapping definitions. Moreover, we omit the \textit{treatment} perspective to avoid specific medical prescriptions and recommendations that might not come from licensed professionals. Eventually, we proceed with the following five perspectives:

\begin{itemize}[noitemsep, nolistsep, topsep=0pt, leftmargin=*]
    \item \textbf{Cause:} It underlines the potential cause of a medical phenomenon or a symptom. It answers the ``Why'' regarding a specific observation, offering insights to identify the root cause. 
    \item \textbf{Suggestion:} It encapsulates strategies, recommendations, or potential courses of action towards management or resolution of a health condition.

    \item \textbf{Experience:} It covers first-hand experiences, observations, insights, or opinions derived from treatment or medication related to a particular problem. 
    
    \item \textbf{Question:} It consists of interrogative phrases, follow-up questions and rhetorical questions that are sought to better understand the context. They typically start with phrases like \textit{Why, What, Do, How}, and \textit{Did} etc, and end in a question mark. 

    \item \textbf{Information:} It encompasses segments that offer factual knowledge or information considering the given query. These segments provide comprehensive details on diagnoses, symptoms, or general information on a medical condition. 
\end{itemize}

\paragraph{Step 1 -- Perspective and Span Annotation:} For a question under consideration, all answers are analyzed for potential perspective labels -- one answer may convey multiple perspectives. Next, the textual span that reflects a particular perspective is marked.


\paragraph{Step 2 -- Summary Annotation:}
Following the perspective and span annotation, summaries are written for each of the identified perspectives. These summaries are a concise representation of the underlying perspective contained within the spans across all answers.
The annotation guidelines can be found in Appendix (\S \ref{subsec:annotation-guidelines}).


\subsection{Annotation Process} We employ three annotators\footnote{The annotators included a master's student, a research assistant, and a research volunteer who is a native English speaker. All the annotators possess reading, writing, and speaking fluency in English.} for annotating the perspectives and summaries. At first, we conduct training sessions for our annotators to familiarize themselves with the annotation guidelines. We also conduct multiple rounds of pilot annotations on a sample size of 50 instances to ensure conformity of the guidelines. Subsequently, we ask our annotators to complete the remaining annotations. 

Finally, we evaluate the annotations via inter-rater agreement scores. For spans, we compute average F1 score (0.88) and average Jaccard similarity(0.85) scores across examples. The average F1 score establishes the agreement over the presence of a span with a particular perspective, and the Jaccard index helps ensure each class's coverage across spans. 
Moreover, we calculate the ROUGE scores \cite{lin-2004-rouge} with R-1, R-2, and R-L values at 0.36, 0.13, and 0.27  to capture the n-gram level similarity between the summaries and BERTScore (0.82) \cite{bert-score} to measure the semantic similarity. 

\begin{figure*}
    \centering
    \includegraphics[width=0.9\textwidth]{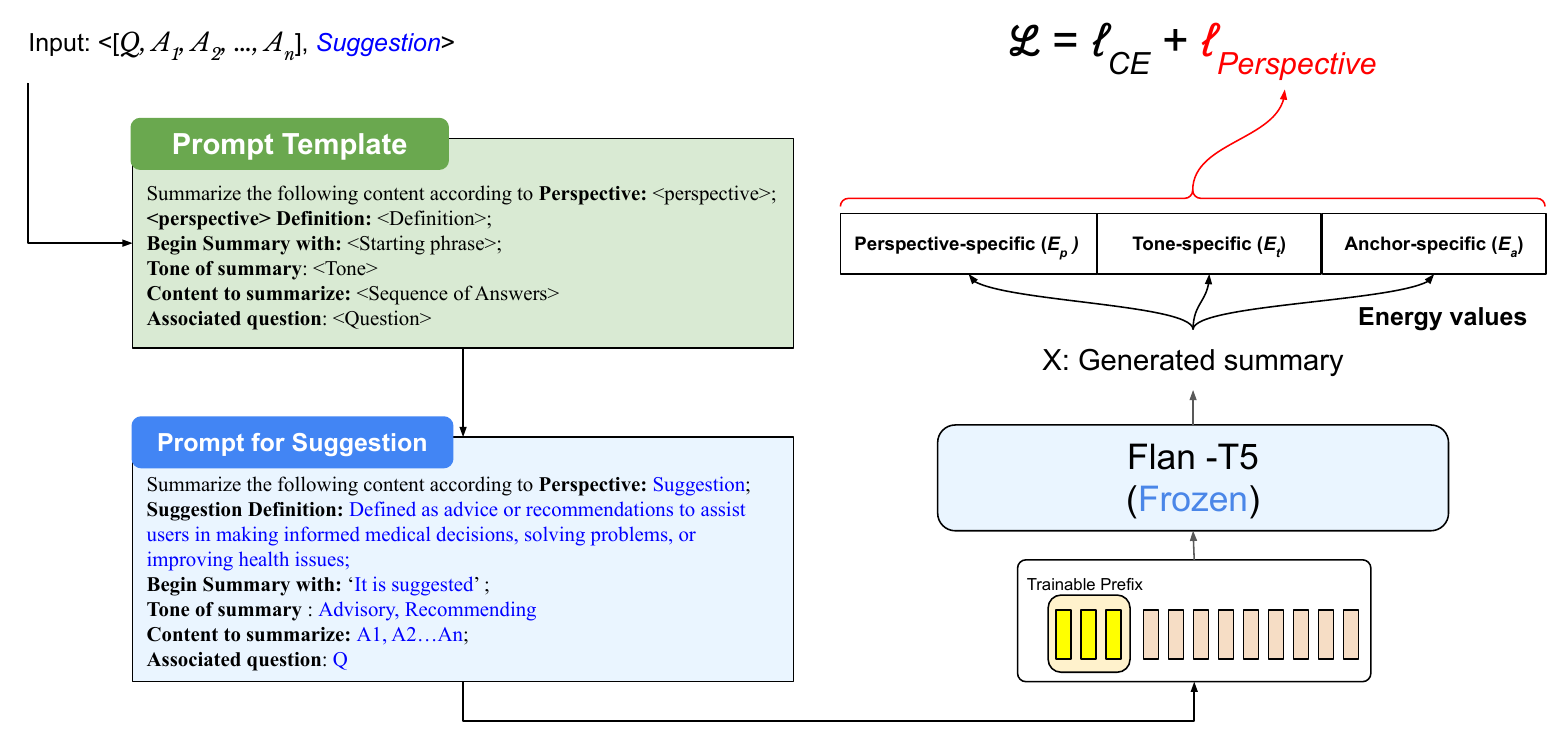}
    \vspace{-3mm}
    \caption{The proposed \model\ model. Given an input, a perspective-conditioned prompt is generated following the prompt template. Subsequently, the prompt is fed to Flan-T5 with prefix tuner to generate the summary. An energy-driven loss function ($\ell_{\text{\em Perspective}}$) is incorporated along with the standard cross-entropy (CE) loss to enforce the perspective attributes in the generated summary.}
    \label{fig:fig_2}
\end{figure*}

\subsection{Data Statistics}


\input{data_stat}

\data\ contains 9987 answer instances for $3167$ questions. We split the dataset into $2533$, $317$, and $317$ question instances for training, validation, and testing, respectively, as shown in Table \ref{tab:data_stat}. We observe that the counts of \textit{information} and \textit{suggestion} are the highest, followed by the \textit{experience} prespective. This is typical to the usage patterns on such CQA forums.  Across the different categories, we find that \textit{suggestion} and \textit{information} perspectives are generally more represented than the other three prespectives. 
 
Although, the distribution of span and summaries across perspectives depicts the richness of the data, it is important to ascertain the healthcare category-wise coverage in the dataset. Figure \ref{fig:fig_cat_sum_span_dist} describes the coverage of spans and summaries, respectively across 17 categories (\S ~\ref{subsec:categories}) for the entire dataset. In case of STDs and infectious diseases, most of the spans were of type \textit{information}.

\section{Methodology}
\label{sec:method}
In this section, we systematically outline the proposed architecture, \model, to generate perspective-specific summaries, depicted in Figure \ref{fig:fig_2}. Given an input -- a question $Q$, a set of answers $A_1, A_2, \dots, A_n$, and a desired perspective $P$, we design a prompt to be fed to \model\ for generating the perspective-controlled summary $S^P$. \model\ incorporates Flan-T5 as the foundational model clubbed with a prefix tuner that adapts the model to the intricacies of the medical perspective summarization. Further, we optimize the architecture through a combination of an energy-controlled domain-specific loss function ($\ell_{\text{\em Perspective}}$) and a standard cross-entropy loss function ($\ell_{\text{\em CE}}$).


\begin{table*}[ht]
\centering
\resizebox{\textwidth}{!}{
\begin{tabular}{lllp{11cm}}
\toprule
\textbf{Perspective} & \textbf{Begin Summary With} & \textbf{Tone}& \textbf{Definition} \\
& \bf (aka. anchor-text) & & \\
\toprule
Information & For information purposes...&Informative, Educational& Defined as knowledge about diseases, disorders, and health-related facts, providing insights into symptoms and diagnosis.
 \\
Cause & Some of the causes...&Explanatory, Causal& Defined as reasons responsible for the occurrence of a particular medical condition, symptom, or disease \\
Suggestion & It is suggested...&Advisory, Recommending& Defined as advice or recommendations to assist users in making informed medical decisions, solving problems, or improving health issues.  \\
Experience & In user's experience...&Personal, Narrative& Defined as individual experiences, anecdotes, or firsthand insights related to health, medical treatments, medication usage, and coping strategies.
 \\
Question & It is inquired...& Seeking Understanding& Defined as inquiry made for deeper understanding.
\\
\bottomrule
\end{tabular}}
\caption{Perspective-specific prompt conditions to design prompts.}
\label{tab:hard_prompt}
\end{table*}

\subsection{Prompt Design}
Effectively, prompt design is an essential component of \model. We carefully design it to refine extensive medical question-and-answer data into focused and perspective-controlled summaries \citep{ravaut2023promptsum} and to specify the relevant perspective nuances and the summary's structure. For instance, summaries focusing on `\textit{experience}' need a narrative and personal tone, whereas ``\textit{suggestion}'' summary is framed in a more advisory and recommending tone. This highlights the need for tailored prompts to guide the model in generating summaries that align with the desired perspective.  

To achieve this for different perspectives, we frame a prompt template structure (as depicted in Figure \ref{fig:fig_2}) using the following components: \textit{Task Detail}, \textit{Perspective Definition}, \textit{Begin Summary with}, \textit{Tone of Summary}, \textit{Content to Summarize}, and \textit{Associated Question}. We supplement these heads with  the input $<[Q, A_1, A_2, \dots, A_n], P>$ and perspective-specific conditions.  



\begin{itemize}[noitemsep, nolistsep, topsep=0pt, leftmargin=*]
    \item \textbf{Task:} This specifies the intended task to perform by \model, e.g., ``\textit{Summarize following content according to perspective: Suggestion}".
    
    \item \textbf{Perspective Definition:} It defines the semantic of the particular perspective that helps the model understand the specific medical context and nuances of the perspective. 
    
    \item \textbf{Begin Summary With:} This prompts guides the model to begin the summary with a specific phrase tailored for the chosen perspective. The starting phrase acts as an initial anchor and hence, directing the model to craft the remainder of the summary with a conscious orientation toward the intended perspective.
    
    \item \textbf{Tone of Summary:} The tone reflects the stylistic approach the summary should take. 
    
    \item \textbf{Content to Summarize:} The answers $[A_1, \dots, A_n]$ is the input document that needs to summarize for the particular perspective.
    
    \item \textbf{Associated Question:} The original question $Q$ associated with the answers provides essential background for the model during summarization.
\end{itemize}

\noindent Table \ref{tab:hard_prompt} outlines perspective-specific conditions for each perspective.

\subsection{Prefix Tuning} 
Inspired by \newcite{li2021prefixtuning}, we adapt prefix tuning to facilitate perspective-specific summary generation in \model. This method involves appending a learnable sequence of continuous vectors, known as ``prefix", to the input of our pre-trained model, Flan-T5. We keep Flan-T5 in a frozen state throughout the process and only train the prefix vectors during the training phase. Consequently, the tuned prefix vectors capture the perspective-specific information and enable the model to tailor summaries in a specific manner according to the input prompt. A significant advantage of prefix-tuning over fine-tuning is its efficiency in parameter utilization -- since only a few parameters are updated to cater to the task rather than updating the entire model parameters. 


\subsection{Energy Controlled Perspective loss}
We develop a controlled perspective loss to explicitly enforce that the generated summary satisfies each constraint. It is inspired by the Energy-Based Models (EBMs) framework, as highlighted in the works of \newcite{mireshghallah-etal-2022-mix} and \newcite{qin2022cold}. EBMs are based on the principles of statistical physics that suggest lower energy values to be more favorable configurations. We apply this principle to compute three energy values considering the perspective-specific ($E_p$), the tone-specific ($E_t$), and the textual anchor-specific ($E_a$). Together, they facilitate the model to align with perspective-specific summary generation and to ensure input prompt conditions.


Given a generated summary $S^P$, the energy value against each perspective $i \in$ \{\textbf{c}ause, \textbf{s}uggestion, i\textbf{n}formation, \textbf{e}xperience, \textbf{q}uestion\} is defined as: 
\begin{equation}
E(S^P)_i = \alpha_1 E_{p_i} + \alpha_2 E_{a_i} + \alpha_3 E_{t_i} \nonumber
\end{equation}

where $\alpha$'s are the hyperparameters.


\begin{itemize}[noitemsep, nolistsep, topsep=0pt, leftmargin=*]
  \item \textbf{Perspective-specific energy value ($E_{p}$):} This defines the probability of a specific perspective $i$ given the input summary $S^P$. To obtain the probability distribution over perspectives, we learn a RoBERTa-based perspective classification model on gold perspective-specific input spans. 
    \begin{equation}
        [E_{p_{c}}, E_{p_{s}}, \dots, E_{p_{q}}] = \text{softmax}(\text{RoBERTa}(S^P)) \nonumber
    \end{equation}
  \item \textbf{Anchor-specific energy value ($E_{a}$):} This evaluates how well the beginning of the generated summary matches the expected anchor-text for the given perspective. It is determined by calculating the Rouge-1 score between the anchor-text of all perspectives (c.f. `\textit{Begin summary with}' in Table \ref{tab:hard_prompt}) and the starting $j$ tokens of the generated summary $S^P$, where $j=len(\text{anchor-text})$. 
    \begin{equation}
        [E_{a_{c}}, E_{a_{s}}, \dots, E_{a_{q}}] = [R1_{{c}}, R1_{{s}}, \dots, R1_{{q}}] \nonumber
    \end{equation}
    
    \item \textbf{Tone-specific energy value ($E_{t}$):} This value is determined by calculating the cosine similarity between the BERT embeddings of generated summary and tone-specific keywords ($k$). For ensuring the semantic coverage, we additionally considers the synonyms of the keywords as well, i.e., $k = k+\text{synonyms}(k)$.
   \begin{equation}
    E_{t_i} = \frac{\text{BERT}(k_i) \cdot \text{BERT}(S^P)}{\|\text{BERT}(k_i)\| \|\text{BERT}(S^P)\|} \nonumber
   \end{equation}
\end{itemize}

\input{results-summary}

Next, we calculate the energy-based probability distribution $p_i(S^P)$
, across all perspectives as follows:
\begin{equation}
p_i(S^P) = \frac{e^{-\frac{1}{E(S^P)_i}}}{\sum_{j} e^{-\frac{1}{E(S^P)_j}}} \nonumber
\end{equation}

Subsequently, we feed the energy-based probability distribution in the cross-entropy function to the compute the perspective loss, 
\begin{equation}
    \ell_{\text{\em Perspective}} = -\sum_{i} y_i \log(p_i(S^P)) \nonumber
\end{equation}
where $y_i \in \{0,1\}$ is the true perspective label.

Finally, we augment the energy-based perspective loss function with the standard cross-entropy function to compute the overall loss.
\begin{equation}
\mathcal{L} = \ell_{\text{\em CE}} + \ell_{\text{\em Perspective}} \nonumber
\end{equation}



\section{Experiments and Results} 
We benchmark \data\ on multiple state-of-the-art approaches. For comparison, we compute ROUGE (R1, R2, and RL) \cite{lin-2004-rouge}, BLEU \cite{papineni-etal-2002-bleu}, Meteor \cite{banerjee-lavie-2005-meteor}, and BERTScore \cite{bert-score}.

\paragraph{Baselines:}
We employ \textbf{GPT-2} \cite{radford2019language}, \textbf{BART} \cite{lewis2019bart}, \textbf{PEGASUS} \cite{zhang2019pegasus}, \textbf{T5} \cite{raffel2019exploring}, and \textbf{Flan-T5} \cite{chung2022scaling} models for comparative analysis. Moreover, we experimented with two variations based on their fine-tuning approaches and input configurations: a) \textit{FDPer} (Fine-tuned on Document and Perspective); and b) \textit{FDProm} (Fine-tuned on Document and Structured Prompt).

\input{results-perspective}
\input{qualitative}

\paragraph{Result Analysis:} 
Table \ref{tab:model-results} shows the results of the baseline models and our proposed model. 
Among all participating baselines, Flan-T5 records the best performance on average as it outperforms all other comparative systems in majority of the cases -- 5 out of 10 cases in the FDPer setup and 4 out of 10 cases in the FDProm setup. Further, we observe that \model\ surpasses all the baseline models across all evaluation metrics. It reports an increment of $+2.69\%$, {$+6.65\%$} and $+1.84\%$ in ROUGE-L, ROUGE-2, and ROUGE-1 scores, respectively, against the best-performing baseline -- indicating the robust syntactic performance of \model. Furthermore, a substantial increase of $+6.58\%$ in the BLEU score denotes enhanced syntactic alignment with human-written summaries, which is critical for the coherence and fluency of the generated text. In assessing the semantic prowess, \model\ continues to exhibit exceptional performance, with an improvement of $+12\%$ in the METEOR score and $+1.15\%$ in BERTScore.
In comparison  with Flan-T5 (FDProm), \model\ exhibits superior performance with prefix-tuning and energy-based methods since it also employs FLAT-T5 as the foundational model. Thus, we can fairly argue that with prefix tuning and energy-driven loss function, \model\ not only saves computational resources but also generates text that is more aligned with better results against Flan-T5. Perspective-wise results of \model\ is listed in Table \ref{tab:perspective-results}.  

\paragraph{Comparison with GPT-4.}
Table \ref{tab:gpt4_vs_plasma} shows comparison between GPT-4 and \model. Due to resource limitations, we randomly select 50 samples from our dataset and using the same prompt used for \model, we generate the perspective-specific summaries from GPT-4. We notice that GPT-4 significantly outperforms \model. However, in comparison, the \model\ model requires significantly fewer parameters than GPT-4, which is trained on massive datasets and has many more parameters. 

\begin{table}[t]
    \centering
    \resizebox{\columnwidth}{!}
    {
    \begin{tabular}{lcccccccccc}
        \toprule
            \multirow{2}{*}{\bf Models}                 & \multicolumn{2}{c}{\bf ROUGE-1} &  \multicolumn{2}{c}{\bf ROUGE-2} & \multicolumn{2}{c}{\bf ROUGE-L} & \multirow{2}{*}{\bf BS}    & \multirow{2}{*}{\bf MET}    & \multirow{2}{*}{\bf BLEU} \\ \cmidrule{2-7}
            & \bf Recall & \bf F1 & \bf Recall & \bf F1 & \bf Recall & \bf F1 & & &  \\ \midrule
 
            GPT-4     & 49.56 & 29.02 &   18.98 & 9.87 & 45.15 & 25.90 & 0.877 & 0.364 & 0.046  \\ \midrule
            \model  & 25.17 & 23.22 & 9.40  &  7.51 & 23.07 & 21.21  & 0.860 & 0.213 & 0.039 \\ \bottomrule
    \end{tabular}
    }
    \caption{Comparison between \model\ and GPT-4 on 50 randomly selected samples.}
    \label{tab:gpt4_vs_plasma}
\end{table}

\paragraph{Ablation Study.}
Ablation results are furnished in Table \ref{tab:ablation-results}. We observe a decline in the performance on removing the energy-controlled perspective loss, thus suggesting its impact on the perspective-specific summary. Further, we experiment with our input prompt by varying the perspective-specific conditions. We observe that all prompt components (i.e., perspective, its definition, tone, and anchor-text) have positive impact on \model. 
Though the perspective word has a significance, inclusion of its definition further improves the performance. Moreover, the inclusion of only tone or only anchor-text along with the perspective word and its definition introduced some noise, their combined presence led to an improvement in the performance of \model. We also experiment with the position of constraint in the prompt (i.e., at the beginning or the end of the main content) and observe better performance with constraint at the beginning (c.f. Appendix).  
Further in Table \ref{tab:results-ablation-energy-components}, we present the results of our ablation study on the energy components. It illustrates that removing any of these components leads to a noticeable drop in performance across various evaluation metrics. This decline indicates that each of the three energy components -- anchor-specific ($E_a$), tone-specific ($E_t$), and perspective-specific ($E_p$) -- is essential for generating high-quality, perspective-specific summaries using the PLASMA model.



\input{results-ablation}
\input{results-ablation-energy-components}
\input{human_evaluation}


\paragraph{Qualitative Analysis.}
In our qualitative evaluation, we compare the output of \model\ with the best baseline, Flan-T5, in Table \ref{tab:qualitative-analysis}. Further, we explore GPT-4 for our use case in a zero-shot setting. 
In the first case for information perspective, we observe that \model\ generates the summary arguably well as compared to the other two systems. \model's generated summary adheres to the input prompt, i.e., anchor text with informative tone and the desired perspective, whereas Flan-T5 captured the information perspective but didn't capture the anchor text. Comparatively, GPT-4, while providing additional context, tends to include information tangential to the main point, resulting in a less focused summary, which could potentially detract from the user's goal of obtaining a concise and perspective-aligned summary. 
In the second instance, upon considering the question perspective, it is observed that both \model\ and Flan-T5 deviate from the desired question perspective, instead they generate the summary in terms of information perspective. On the other hand, GPT-4 captures the anchor text of the question perspective but continues elaborating in an informative way. Our analysis suggests that \model\ and the rest of the baselines perform poorly with question perspective, possibly due to relatively fewer samples.

\paragraph{Human Evaluation.}
We conduct a comprehensive human evaluation on a random subset of 25 threads evaluated by 50 participants to assess the quality of summaries generated by our proposed method \model\, the best baseline, FlanT5, and GPT-4 against the reference summaries. We perform two-stage assessments: \textit{perspective identification} and \textit{qualitative summary assessment}. In perspective identification, each participant was first presented with the anonymized summaries along with the input document and was asked to identify the perspective they believed was represented in the summary. We calculated the \textit{perspective accuracy} for each case based on the feedback. Following the perspective identification, participants were informed of the actual perspective intended for the summary and, subsequently, were asked to assess the quality of summaries based on six criteria - \textit{fluency}, \textit{coherence}, \textit{consistency}, \textit{extractiveness}, \textit{capturing perspective}, and \textit{faithfulness} -- on a Likert scale of 1-5. We define these parameters as follows: 
\begin{itemize}
\item \textit{Fluency}: Assesses how easily the text can be read and understood, checking for grammatical and syntactic correctness.

\item \textit{Coherence}: Evaluates the logical flow and clarity, ensuring well-connected sentences form a coherent narrative.

\item \textit{Consistency}: Verifies factual accuracy and alignment with the source content, ensuring no discrepancies or distortions.

\item \textit{Capturing Perspective}: Determines if the summary accurately reflects the intended perspective.

\item \textit{Extractiveness}: Measures the proportion of information directly copied from the original post.

\item \textit{Faithfulness}: The degree to which the information in the summary stays true to the original text's facts, assertions, and general intent. 
\end{itemize}
We compute the mean of all scores and present them in Table \ref{tab:human_eval}. 
Our evaluation shows that \model\ surpasses the best baseline, FlanT5, in all assessed metrics and outperforms GPT-4 in several key areas. Specifically, \model\ demonstrates superior performance over GPT-4 in terms of fluency, consistency, extractiveness, and faithfulness.

\section{Conclusion}

In response to the lack of perspective-specific summarization datasets in healthcare, we introduced \data\, a pioneering dataset specifically designed for the perspective-specific summarization task. Our dataset features five perspective labels (Suggestion, Information, Question, Cause, and Experience). Further, to benchmark the dataset, we propose \model\, which incorporates efficient customization of the model's behavior based on the input prompt without the need for extensive retraining and energy-based loss to cater to a custom-designed multi-attribute input prompt. We conducted an extensive evaluation (i.e., empirical, qualitative) to establish the effectiveness of the \model.

\section{Limitations}
Building an accurate model using vetted medical data is challenging due to the sensitivity of the domain. As an alternative, leveraging healthcare CQA forums offers a vast resource, but regular LLM-based approaches compromise factuality and accountability. We made our best efforts with prompt designing and controlling perspective-guided summarization. However, we expect forums to often include subjective opinions, marketing content, and various other forms of noise that can bring natural bias. Our baseline models often struggled with co-occurring perspectives that lacked specific patterns. For instance, \textit{Information} lacked clear speech indicators, while \textit{Suggestions} featured directive phrases like "should see" or "is recommended" \textit{Questions} typically ended with a question mark, and \textit{Experiences} frequently included personal and first-person singular pronouns. One of the limitations of the dataset is the imbalanced number of samples, which hampers the generation, as seen in the question perspective. In managing the potential risks associated with disseminating community-sourced medical advice through our summarization model, a key decision was to exclude a distinct 'treatments' perspective. This choice was driven by ethical considerations aimed at minimizing the risk of our tool being perceived as a source of medical advice or endorsing specific treatments. However, it is recognized that other perspectives like 'suggestion,' 'cause,' and 'information' may still indirectly convey medical advice, reflecting the broader challenges within CQA forums where personal experiences, factual information, and speculative advice merge.



\section{Ethical Considerations}
Since our dataset pertains to the medical and healthcare domain, we have committed to withstand the standard ethical practices \cite{bear2022scoping, fu2023recommended}. Given that the dataset is obtained from Yahoo! Answers Corpus, a social media platform, a risk of revealing highly personal health-related content is assumed to already exist in the public domain. Although user profiles were anonymized, not all identifiable information was removed from the answers, like a clinic's or a doctor's user info, name, etc, considering the user's willingness to reveal it on public platforms. We also observed answers containing sales pitches veiled as medical advice and links to inactive websites. To mitigate these issues, we strictly adhered to annotation guidelines, avoiding the annotation of personal identifiers. Also, no attempts were made to interact or connect to users on their other social media handles, avoiding any risk associated with back-tracing. Furthermore, we used neutral and general terms such as "information, "experience", "suggestion", "cause", and "question" in our naming conventions, avoiding medically complex terms like "treatment" that may get associated, in some cases, confused with specific medical advice. This approach was taken to prevent the dissemination of incorrect medical guidance through our annotations or summaries. Finally, it was emphasized that our annotators were not medically trained, reflecting our aim to extract medical data from a layperson's perspective. Every intellectual artifact and resource was cited to the best of our knowledge. We emphasize that with our efforts in this research, our aim is not to provide any medical solution but instead assist internet users in retrieving essential information easily.

\section*{Acknowledgment}
We would like to acknowledge Ghadir Alsahmi for her support in annotation and evaluation. Md Shad Akhtar would like to acknowledge the support of Infosys foundation through Center of AI (CAI)-IIIT Delhi.

\bibliography{acl_latex} 

\appendix
\section{Appendix}
\label{sec:appendix}
\subsection{Annotation Guidelines}  
\label{subsec:annotation-guidelines}
Based on the definitions provided in section \ref{annotation_process}, the annotators are instructed to follow the following instructions :
\begin{enumerate}[leftmargin=*]
    \item Validate the document's alignment with the medical domain, ensuring content pertains to one or more of the following health-related topics: prevention, diagnosis, management, treatment of diseases, understanding of bodily functions or processes, the effects of medications or medical interventions, and queries regarding wellness practices.
    \item  Assess each text span within the context of the given post or topic and select the most relevant perspective that adheres to the perspective definitions \ref{annotation_process}
    \item Avoid personal bias when assigning perspectives to text spans
    \item Multi-perspective labeling is allowed for a span of text 
    \item Do not annotate any links and personally identifiable text provided in the input document
    \item Assign the relevant perspective label to a segment of text where the quantity of medicines or the duration of medicine ingestion is explicitly mentioned in the text.
    \item Review the classified spans again not to miss any underlying perspective.
    \item While writing summaries, carefully understand the extracted spans to capture essential ideas and significant medical details from the text and concerning the perspective of the annotated spans 
    \item Create concise summaries molded according to the essence of each perspective.
    \item Frame summaries appropriately:
    \begin{itemize}[leftmargin=*]
        \item For Information perspective summaries, initiate with phrases like "For information purposes."
        \item For Suggestion perspective summaries, initiate with phrases like "It is suggested," "It is advised," or "Consider."
        \item For Experience perspective summaries, commence with phrases like "One user shared his experience" or "In user's experience."
        \item For Cause perspective summaries, initiate with phrases like "Some of the causes".
        \item For Question perspective summaries, initiate with phrases addressing inquiries directed to the questioner, such as "It is inquired."
    \end{itemize}
    \item Refrain from adding any additional information in the summaries beyond what is explicitly provided in the document. 
\end{enumerate}
\subsection{Disease Categories}\label{subsec:categories}
    Infectious Diseases, Women's Health, STDs, Mental Health, Heart Diseases, Other - Health, Skin Conditions, First Aid, Diabetes, Allergies, Dental, Cancer, Men's Health, Diet \& Fitness, Respiratory Diseases, Alternative Medicine, Other - Diseases

\subsection{Annotation Tool Development}\label{subsec:brat}
As described in the earlier sections , we make use of \textbf{B.R.A.T} v1.4 for the annotations. We employ this tool to label spans of text with perspectives. As BRAT is generally used with structured notes and not freeform text, we transform our data from a JSON to a text file for BRAT to be able to work. Additionally, our task involves grouping similar spans of text based on their \textit{perspective}. So we build two features into BRAT. (a) Group spans groups all spans of labeled text by their perspective type and joins them (b) Delete Groups - Delete groups is provided to revert to the state prior to grouping if an annotator decides to make changes to the spans. 


\begin{figure}
    \centering
    \includegraphics[width=1\linewidth]{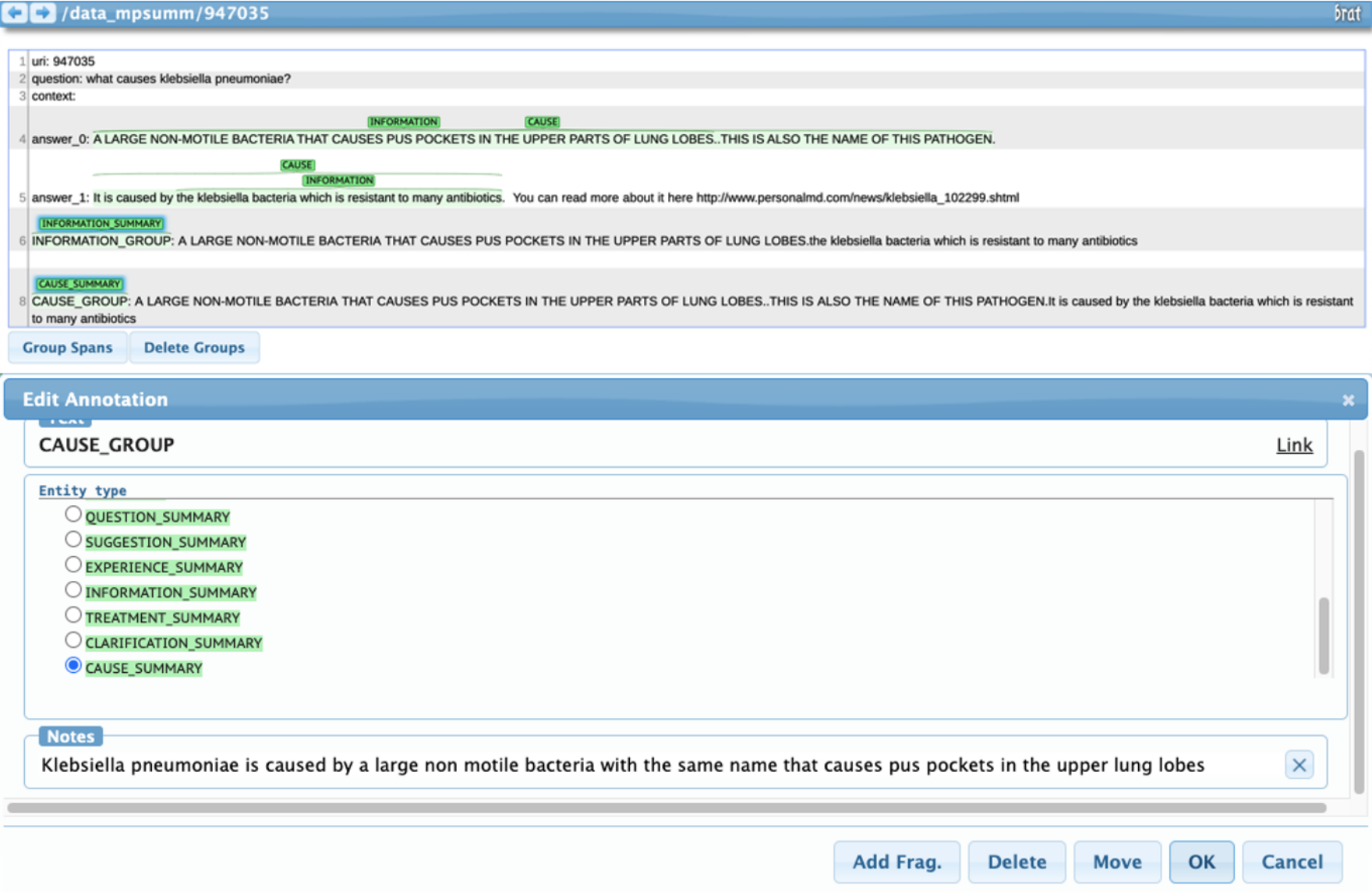}
    \caption{Labelling spans and Writing perspective oriented summaries}
    \label{fig:fig_4}
\end{figure}


An annotator labels the span as per the annotation guidelines described in Section \ref{sssec:ann_scheme}. After annotating the span, there are two choices, i.e., to edit the spans or to group them. After grouping, perspective-oriented summary are written in the notes section as described in Figure \ref{fig:fig_4}. 

\subsection{Dataset Statistics}
These figures illustrate some statistics from \data\
\begin{figure}[htbp]
    \includegraphics[width=\linewidth] {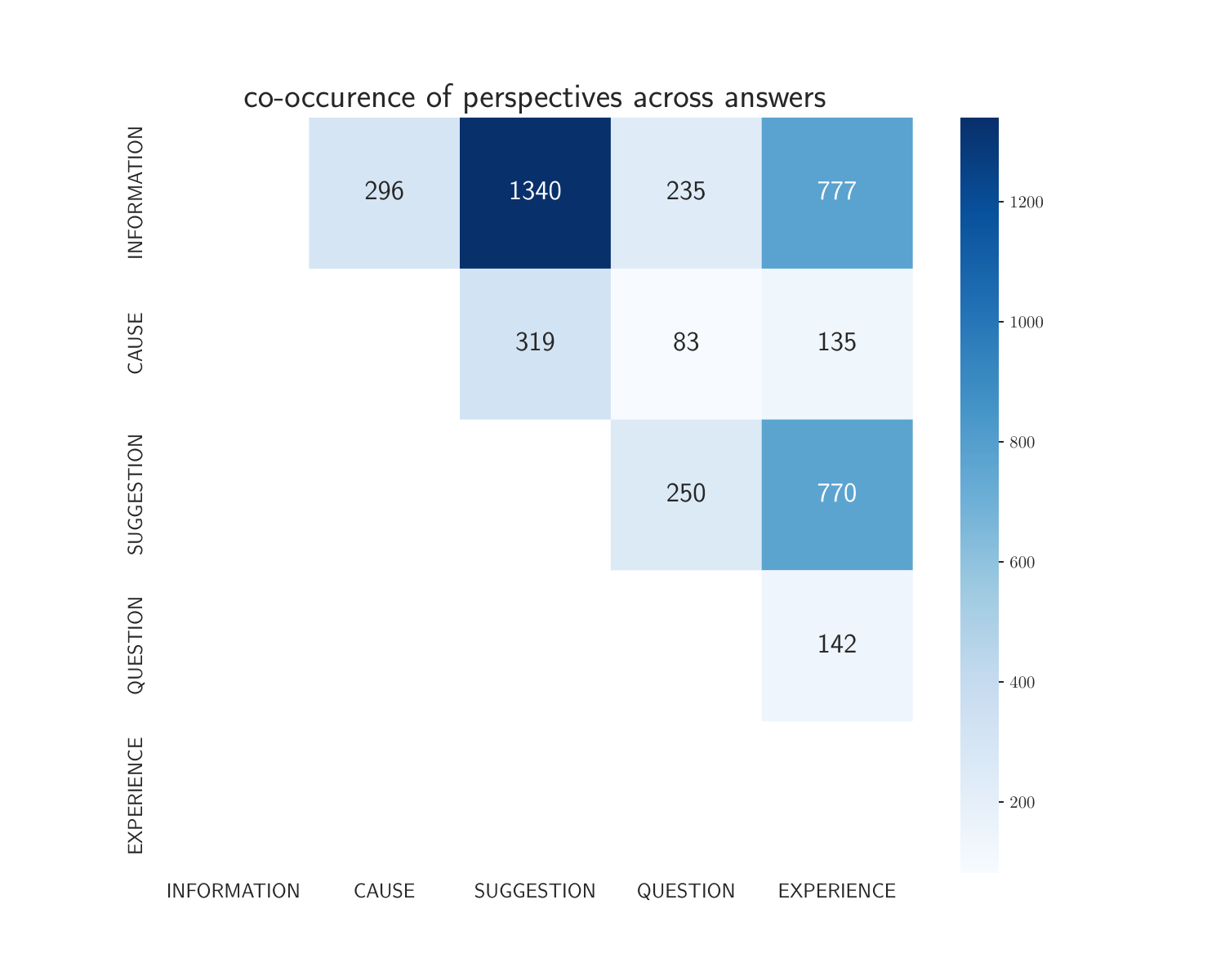}
    
    \caption{Perspective co-occurence across answers}
    \label{fig:fig_cooc1}
\end{figure}
\begin{figure}[htbp]
    \includegraphics[width=\linewidth]{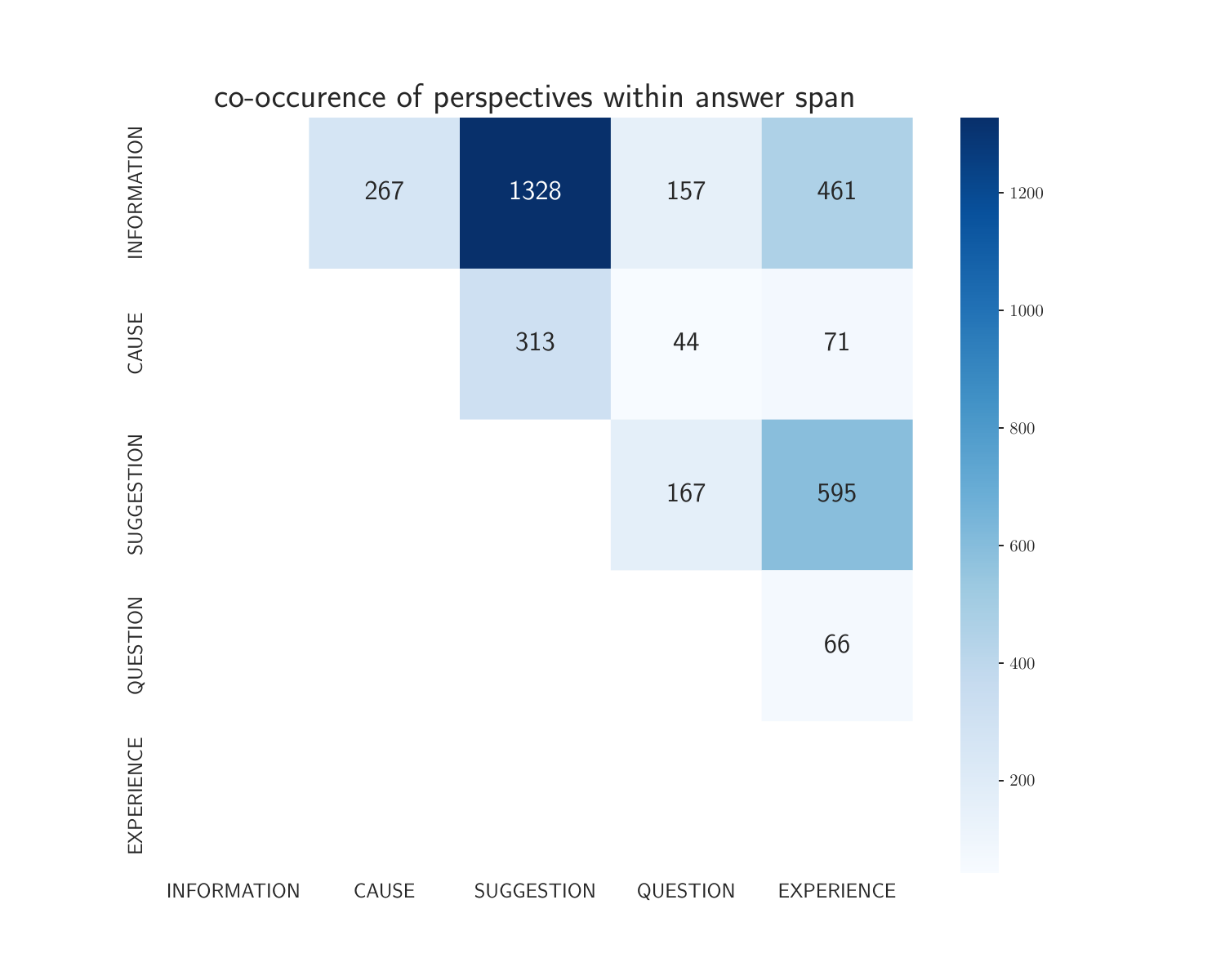}
    \caption{Perspective co-occurence within answer span}
    \label{fig:fig_cooc2}
\end{figure}

\subsection{Additional analysis on the dataset}
\input{span_tables}
 From Figure \ref{fig:fig_cooc1} and \ref{fig:fig_cooc2}, we can see the co-occurrence of perspectives across answers and across selected spans of an answer. The first plot goes to show the co-occurrence of the perspectives across different answers. The second plot however plots the co-occurence of perspectives for a given answer to a question, hence showing the different perspectives embedded within an answer. From the second plot we find that often the pairs \textit{Information-Suggestion}, \textit{Information-Experience} and \textit{Suggestion-Experience} occur together the most. An example interpretation would be, "given an answer, it is more likely to observe \textit{Information} and \textit{Suggestions} together". These plots also confirm the hypothesis we started with that in the biomedical domain, there are well defined perspectives, unlike the open domain, as there is some co-occurrence, but Table \ref{tab:data_stat} shows that most of the spans do convey their own meaning.

\subsection{Analysis of placement of prompt}

Table \ref{tab:placement-prompt} showcases the impact of placing constraints before or after the main content in the prompt when using the PLASMA model. These constraints refer to perspective-specific attributes designed to enhance the generation of perspective-specific summaries by the Flan-T5 model, as illustrated in Figure \ref{fig:fig_2}. The results demonstrate that placing these constraints before the main content in the prompt significantly improves the model's performance across all metrics thus effectively guiding the model to generate summaries that more accurately reflect the desired attributes and quality.

 \begin{table}[!h]
    \centering
    \resizebox{\columnwidth}{!}
    {
    \begin{tabular}{lcccccccccc}
        \toprule
            \multirow{2}{*}{\bf Models}                 & \multicolumn{2}{c}{\bf ROUGE-1} &  \multicolumn{2}{c}{\bf ROUGE-2} & \multicolumn{2}{c}{\bf ROUGE-L} & \multirow{2}{*}{\bf BS}    & \multirow{2}{*}{\bf MET}    & \multirow{2}{*}{\bf BLEU} \\ \cmidrule{2-7}
            & \bf Recall & \bf F1 & \bf Recall & \bf F1 & \bf Recall & \bf F1 & & &  \\ \midrule
               \model (Placement Before)  & \bf 30.16 & \bf23.23 & \bf 10.23  &  \bf7.38 & \bf 27.78 & \bf21.38  & \bf 0.869 & \bf 0.244 & \bf 0.0405 \\ \midrule
\model (Placement After)     & 23.10 &  20.90 &   7.40 & 5.58 & 21.88 & 19.86 & 0.844 & 0.106 & 0.0144  \\ \bottomrule
    \end{tabular}
    }
    \caption{Comparison between adding constraints before and after \model. }
    \label{tab:placement-prompt}
\end{table}

\end{document}

%% file: data_stat.tex
\begin{table}[!t]
\centering
\small
\resizebox{0.5\textwidth}{!}{
\begin{tabular}{lccccc}
\toprule
& \bf Information & \bf Cause & \bf Suggestion & \bf Question & \bf Experience \\
\cmidrule{1-6}
Train (2533)      & 4823/1961 & 646/342 & 4128/1547 & 325/249 & 1439/845 \\
Validation (317)  & 643/246 & 108/49 & 549/208 & 42/32 & 170/108 \\
Test (317)  & 631/242  & 81/45  & 499/188  & 44/31 & 181/100  \\ \hline
\textbf{Total (3167)} & \textbf{6097/2449} & \textbf{835/436} & \textbf{5176/1943} & \textbf{411/312} & \textbf{1790/1053} \\ \bottomrule
\end{tabular}
}
\caption{Dataset Statistics - each cell describes the perspective specific span count/summaries count in that set}
\label{tab:data_stat}
\end{table}

%% file: results-summary.tex
\begin{table}[t]
\centering
\resizebox{\columnwidth}{!}
{
\begin{tabular}{llcccccccccc}
\toprule
 & \multirow{2}{*}{\bf Models}                 & \multicolumn{2}{c}{\bf ROUGE-1} &  \multicolumn{2}{c}{\bf ROUGE-2} & \multicolumn{2}{c}{\bf ROUGE-L} & \multirow{2}{*}{\bf BS}    & \multirow{2}{*}{\bf MET}    & \multirow{2}{*}{\bf BLEU} \\ \cmidrule{3-8}
 & & \bf Recall & \bf F1 & \bf Recall & \bf F1 & \bf Recall & \bf F1 & & &  \\ \midrule
 \multirow{5}{*}{\rotatebox{90}{FDPer}}
 & FLAN-T5      & 25.34     & 22.81  & 7.83          & 6.03          & 23.23     & 18.8         & 0.854 & 0.210     & 0.036 \\
 & GPT2         & 21.49     & 20.54          & 6.44          & 5.72          & 19.87     & 20.55        & 0.855 & 0.134     & 0.030 \\
 & BART         & 20.18     & 21.26          & 6.79     &    6.92  & 18.35     & 19.34        & 0.865 & 0.171     & 0.032 \\
 & PEGASUS      & 19.65     & 19.23          & 5.44          & 5.55          & 17.31     & 17.01        & 0.851 & 0.159     & 0.027 \\
 & T5           & 19.88     & 22.73          & 6.04          & 6.10          & 20.20     & 19.49        & 0.860 & 0.172     & 0.030 \\ \midrule

\multirow{5}{*}{\rotatebox{90}{FDProm}} 
& FLAN-T5     & 25.90 & 21.22 &  8.28         & 6.50 & 23.80 & 20.82 & 0.852          & 0.217 & 0.034  \\
& GPT2        & 19.81         & 14.86 &  4.86         & 6.60 & 19.19         & 17.79         & 0.834          & 0.148         &  0.023 \\
& BART        & 24.17         & 22.69 &  8.74 & 6.66 & 22.09         & 20.65         & 0.867  & 0.020        & 0.038 \\
& PEGASUS     & 20.32         & 18.52 &  4.77         & 4.53 & 17.72         & 16.17         & 0.849          & 0.1679        &  0.025 \\
& T5          & 11.69         & 13.40 &  4.04         & 3.80 & 11.11         & 12.67         & 0.836          & 0.103         &  0.020 \\

\midrule
\rowcolor{blue!20} & \model  & \bf 30.16 & \bf23.23 & \bf 10.23  &  \bf7.38 & \bf 27.78 & \bf21.38  & \bf 0.869 & \bf 0.244 & \bf 0.0405 \\ \bottomrule
\end{tabular}}
\caption{Comparison between \model\ and baselines. BS and MET refer to BERTScore and METEOR.}
\label{tab:model-results}
\end{table}

%% file: results-perspective.tex
\begin{table}[!t]
\centering
\small
\resizebox{\columnwidth}{!}{
\begin{tabular}{lcccccc}
\toprule
\bf Perspective & \bf R1        & \bf R2       & \bf RL         & \bf BERTScore & \bf METEOR    & \bf BLEU \\ \midrule
Information & 27.68    & 10.54   & 25.66     & 0.859      & 0.178        & 0.030 \\
Suggestion  & 22.56         & 5.66         & 20.60          & 0.859         & 0.171    & 0.022 \\
Experience  & 18.91         & 5.23         & 17.41          & 0.859        & 0.164    &  0.023 \\
Question    & 10.88 & 1.31 & 9.48   & 0.860      & 0.150   & 0.010 \\
Cause       & 21.62         & 7.55         & 20.00          & 0.859  &    0.217  & 0.036  \\ 
\bottomrule
\end{tabular}
}
\caption{Perspective-wise scores for \model\ .}
\label{tab:perspective-results}
\end{table}

%% file: qualitative.tex
\begin{table*}[!]\centering
\resizebox{\textwidth}{!}{
\begin{tabular}{llp{70em}}\toprule
\multirow{10}{*}{\rotatebox{90}{\bf Information}} & \textbf{Gold} & For information purposes, donating blood is generally considered to be a healthy practice, and it is safe to donate within the limits set by blood donation organizations. The average human body has about 5 pints of blood, and blood cells have a lifespan of about 30 days. Donating blood periodically, such as once a year, can contribute to maintaining a fresh and healthy blood supply in the body. \\ \cmidrule{2-3}
& \textbf{Flan-T5} & The average human body has about 5 pints of blood at any given time. Each individual blood cell has a life span of about 30 days from birth to death. Therefore, you have a completely new supply of blood about every 30 days. So, no matter how much blood you donated/lost, you would be back to normal 30 days later. \\ \cmidrule{2-3}
& \textbf{GPT-4} & For information purposes, donating blood is considered beneficial to health, allowing individuals to contribute to saving lives while also detoxifying their body. The human body, which holds about 5 pints of blood, replenishes its blood supply every 30 days, ensuring quick recovery post-donation. However, it's crucial to donate responsibly, as excessive loss can lead to shock, highlighting the importance of moderation.\\  \cmidrule{2-3}
& \textbf{\model} & For information purposes, the average human body has about 5 pints of blood at any given time. Therefore, you have a completely new supply of blood about 30 days from birth to death. So, no matter how much blood you donated/lost, you would be back to normal 30 days later. \\
\midrule \midrule
\multirow{6}{*}{\rotatebox{90}{\bf Question}} & \textbf{Gold} & It is inquired regarding why a 2-year-old would get 12 mg of a methadone by iv at all.
 \\ \cmidrule{2-3}
& \textbf{Flan-T5} & Assuming that this child weighs around 25-30 pounds (11-14 kilograms), a typical STARTING DOSE of IV methadone would be 1.1-1.4mg given 2 to 6 times per day. \\ \cmidrule{2-3}
& \textbf{GPT-4} & It is inquired if a 2-year-old should receive 12 mg of IV methadone, given that standard dosages range from 1.1 to 1.4 mg. Such a high dose could be dangerous without a history of treatment for severe conditions like cancer. This dosage may lead to serious side effects, emphasizing the critical need for medical consultation. The situation underscores the complexities of prescribing methadone for pain in young children and the importance of informed medical guidance. \\ \cmidrule{2-3}
& \textbf{\model} & This would put the child at great risk of serious side effects (i.e. he/she may stop breathing or suffer serious cardiovascular problems). 
 \\  \bottomrule
\end{tabular}}
\caption{Comparative analysis on the quality of generated summaries against the gold summary.}
\label{tab:qualitative-analysis}
\end{table*}

%% file: results-ablation.tex
\begin{table}[t]
\centering
\resizebox{\columnwidth}{!}
{
\begin{tabular}{llcccccc}
\toprule
\multicolumn{2}{l}{\bf Models} & \bf R1 & \bf R2 & \bf RL & \bf BERTScore & \bf METEOR & \bf BLEU \\ \midrule
\rowcolor{blue!20} \multicolumn{2}{l}{\model}  & \bf23.23   &  \bf7.38 & \bf21.38 & \bf 0.869 & \bf 0.244 & \bf 0.0405 \\ \midrule

& $- \ell_{\text{\em Perspective}}$ & 22.22 & 6.80 & 20.37 & 0.8518 & 0.223 &  0.033 \\ \midrule
\multirow{5}{*}{\rotatebox{90}{Prompt}} & $P$   & 17.46 & 5.60 & 16.05 & 0.847 & 0.154 &  0.027 \\
& $D$     & 20.11 & 6.18 & 18.52 & 0.849 & 0.188 &  0.028\\
& $P+D$   & 21.22 & 6.58 & 19.60 & 0.857 & 0.223 &  0.034\\
& $P+D+B$ & 19.81 & 6.00 & 18.18 & 0.846 & 0.183 &  0.027\\
& $P+D+T$ & 19.99 & 6.27 & 18.30 & 0.845 & 0.187 &  0.030\\
\bottomrule
\end{tabular}
}
\caption{Ablation results on \model. Prompts are framed using perspective-specific rules: ($P$)erspective, ($D$)efinition, ($T$)one, \& ($B$)egin summary.}
\label{tab:ablation-results}
\end{table}

%% file: results-ablation-energy-components.tex
\begin{table}[t]
\centering
\resizebox{\columnwidth}{!}{%
\begin{tabular}{llcccccc}
\toprule
\multicolumn{2}{l}{\bf Models} & \bf R1 & \bf R2 & \bf RL & \bf BERTScore & \bf METEOR & \bf BLEU \\ \midrule
\rowcolor{blue!20} \multicolumn{2}{l}{\model}  & \bf23.23 & \bf7.38 & \bf21.38 & \bf 0.869 & \bf 0.244 & \bf 0.0405 \\ \midrule

\multirow{3}{*}{\rotatebox{90}{EC}} 
& $ - E_a$     & 16.57 & 15.29 & 5.33 & 0.850 & 0.148 &  0.0194\\
& $ - E_t$     & 17.95 & 16.399 & 5.79 & 0.849 & 0.1495 &  0.0195\\
& $ - E_p$     & 20.22 & 18.72 & 7.26 & 0.845 & 0.1736 &  0.0267\\
\bottomrule
\end{tabular}
}
\caption{Ablation results on Energy Components (EC) of Energy Controlled Perspective loss where $E_a$ is Anchor-specific energy value, $E_t$ is Tone-specific energy value, and $E_p$ is Perspective-specific energy value.}
\label{tab:results-ablation-energy-components}
\end{table}

%% file: human_evaluation.tex
\begin{table*}[th]
\centering
\resizebox{\textwidth}{!}{%
\begin{tabular}{lccccccc}
\toprule
\bf Summary Types & \bf Perspective Accuracy(\%) & \bf Fluency & \bf Coherence & \bf Consistency & \bf Extractiveness & \bf Capturing Perspective & \bf Faithfulness \\ \midrule
\rowcolor{green!20} Reference & 92.65 & 4.42 & 4.29 & 4.21 & 4.10 & 4.53 & 4.75 \\ \midrule
PLASMA & 87.27 & \bf 3.83 & 3.76 & \bf 3.62 & \bf 3.55 & 3.89 & \bf 3.98 \\ \midrule
Flan-T5 & 71.25 & 3.39 & 3.70 & 3.40 & 3.48 & 3.76 & 3.81 \\
\midrule
GPT-4 & \bf 93.56 & 3.63 & \bf 3.88 & 3.55 & 3.38 & \bf 3.95 & 3.66 \\
\bottomrule
\end{tabular}
}

\caption{Human evaluation on 25 threads evaluated by 50 participants.}
\label{tab:human_eval}
\end{table*}